\documentclass{bmvc2k}
\usepackage{bm, multirow, amsmath, amssymb, array, floatrow, subcaption}

\title{Synthetic Training for Accurate 3D Human Pose and Shape Estimation in the Wild}

\addauthor{Akash Sengupta}{as2562@cam.ac.uk}{1}
\addauthor{Ignas Budvytis}{ib255@cam.ac.uk}{1}
\addauthor{Roberto Cipolla}{rc10001@cam.ac.uk}{1}

\addinstitution{
 Department of Engineering\\
 University of Cambridge\\
 Cambridge, UK
}

\runninghead{Sengupta et al.}{Synthetic Training for 3D Human Pose and Shape}

\def\eg{\emph{e.g}\bmvaOneDot}

\def\etal{\emph{et al}\bmvaOneDot}
\newfloatcommand{capbtabbox}{table}[][\FBwidth]
\DeclareCaptionLabelFormat{andtable}{#1~#2  \&  \tablename~\thetable}

\begin{document}
\maketitle
\begin{abstract}
This paper addresses the problem of monocular 3D human shape and pose estimation from an RGB image. Despite great progress in this field in terms of pose prediction accuracy, state-of-the-art methods often predict inaccurate body shapes. We suggest that this is primarily due to the scarcity of \textit{in-the-wild} training data with \textit{diverse and accurate} body shape labels. Thus, we propose STRAPS (Synthetic Training for Real Accurate Pose and Shape), a system that utilises proxy representations, such as silhouettes and 2D joints, as inputs to a shape and pose regression neural network, which is trained with synthetic training data (generated on-the-fly during training using the SMPL statistical body model) to overcome data scarcity. We bridge the gap between synthetic training inputs and noisy real inputs, which are predicted by keypoint detection and segmentation CNNs at test-time, by using data augmentation and corruption during training. In order to evaluate our approach, we curate and provide a challenging evaluation dataset for monocular human shape estimation, Sports Shape and Pose 3D (SSP-3D). It consists of RGB images of tightly-clothed sports-persons with a variety of body shapes and corresponding pseudo-ground-truth SMPL shape and pose parameters, obtained via multi-frame optimisation. We show that STRAPS outperforms other state-of-the-art methods on SSP-3D in terms of shape prediction accuracy, while remaining competitive with the state-of-the-art on pose-centric datasets and metrics.
\end{abstract}

\section{Introduction}
\label{sec:intro}

\begin{figure}[t]
    \centering
    \includegraphics[width=\textwidth]{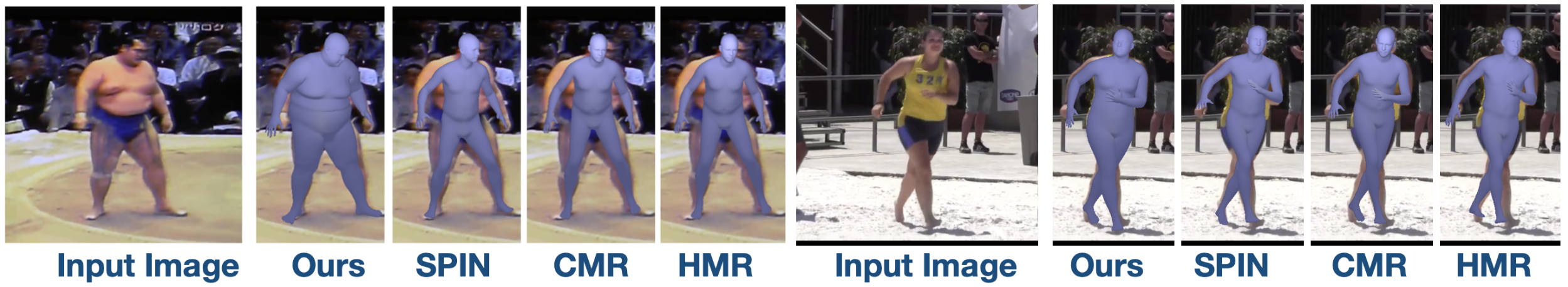}
    \caption{\textbf{STRAPS predicts body shapes with greater accuracy than other approaches} to monocular human 3D shape and pose estimation, such as SPIN \cite{kolotouros2019spin}, CMR \cite{kolotouros2019cmr} and HMR \cite{hmrKanazawa17}, without requiring training images annotated with 3D labels. The images shown in this figure are part of the dataset we provide, SSP-3D.}
    \label{fig:intro_fig}
\end{figure}

3D human shape and pose estimation from a single RGB image is a challenging computer vision problem, with widespread applications in computer animation and augmented reality. Recently, several deep-learning-based methods have been proposed \cite{tan2017, pavlakos2018humanshape, hmrKanazawa17, omran2018nbf, kolotouros2019spin, kolotouros2019cmr, zhang2019danet, varol18_bodynet, Xu_2019_ICCV}. Such methods provide impressive 3D pose reconstructions given single RGB images as inputs, by leveraging datasets of images of humans in a diverse range of labelled 3D poses \cite{h36m_pami, varol17_surreal, vonMarcard2018, movi2020, Lassner:UP:2017, mono-3dhp2017}. However, these approaches often predict inaccurate body shapes, as shown in Figure \ref{fig:intro_fig}. We suggest that this is due to a lack of body shape diversity within the prevalent training datasets. Most learning-based models will struggle to generalise to unseen test data if the distribution of the test data is significantly different from the training data distribution. Thus, increasingly inaccurate body shapes are predicted as the shape of the test subject is further removed from the training datasets' mean shape.

In this paper, we present \textbf{S}ynthetic \textbf{T}raining for \textbf{R}eal \textbf{A}ccurate \textbf{P}ose and \textbf{S}hape (STRAPS), a deep-learning-based framework that uses synthetic training data to overcome the lack of shape diversity in current datasets. Given an input image, inference occurs in two stages (see Figure \ref{fig:straps}). First, we predict a proxy representation, which encodes the subject's silhouette and 2D joint locations, using off-the-shelf segmentation and 2D keypoint detection CNNs \cite{wu2019detectron2, he2017maskrcnn, Guler2018DensePose}. Then, we use a neural network regressor to predict the parameters of a statistical body model (SMPL \cite{SMPL:2015}) from the proxy representation. The regressor is trained using synthetic input-target pairs generated on-the-fly during model training. This is done by sampling target SMPL shape and pose parameters from a training distribution and rendering the corresponding silhouettes and 2D joints, which act as inputs. Since we can choose the form of the training distribution, we have control over the diversity of human body shapes seen by the regressor during training. We utilise simple data augmentation and corruption techniques (see Figure \ref{fig:straps}) to make our regressor robust to noisy inputs encountered at test-time.

Moreover, we curate and provide Sports Shape and Pose 3D (SSP-3D), a dataset which contains images of tightly-clothed sports-persons with a diverse range of body shapes in varied environments, obtained from the Sports-1M video dataset \cite{KarpathyCVPR14}. We use multi-frame optimisation, with forced shape consistency between frames, to obtain pseudo-ground-truth SMPL shape and pose parameters for the sports-person in each image. We evaluate our neural network regressor, along with several recent learning-based approaches, on SSP-3D and report shape prediction accuracy in terms of per-vertex Euclidean error in a neutral pose. Examples from SSP-3D are shown in Figure \ref{fig:ssp3d_samples_statistics}, along with statistics illustrating the greater body shape diversity in SSP-3D compared to widely-used 3D human datasets.

In summary, we have two main contributions: (i) a deep-learning framework which uses synthetic training data and simple data augmentation techniques to overcome the lack of body shape diversity within prevalent datasets and (ii) the SSP-3D evaluation dataset, which we use to show that our neural network regressor results in better shape prediction accuracy than other competing approaches. Our code and dataset are available for research purposes at \url{https://github.com/akashsengupta1997/STRAPS-3DHumanShapePose}.

\section{Related Work}
\label{sec:related_work}

In this section, we discuss recent approaches to 3D human pose and shape estimation, as well as the training datasets typically used for this task.

\noindent \textbf{Monocular 3D human pose and shape estimation} approaches can be classified into two paradigms: optimisation-based and learning-based. Optimisation-based approaches attempt to fit a parametric body model \cite{Anguelov05scape:shape, SMPL:2015, SMPL-X:2019} to 2D observations, such as 2D joints \cite{Bogo:ECCV:2016, SMPL-X:2019}, body surface landmarks \cite{Lassner:UP:2017}, silhouettes \cite{Lassner:UP:2017} or body part segmentations \cite{Zanfir_2018_CVPR}. These approaches produce reliable results without requiring 3D-labelled datasets, which are expensive to obtain. However, they are slow at test-time, sensitive to initialisation and can get stuck in bad local minima, which motivates learning-based approaches.

Learning-based approaches can be further divided into two types: non-parametric 3D regression and body model parameter regression. Non-parametric 3D regression involves predicting a 3D human body representation from an image, such as a voxel occupancy grid \cite{varol18_bodynet} or vertex mesh \cite{kolotouros2019cmr}. However, each representation has associated drawbacks for body shape prediction: \eg voxels are limited by the resolution of the voxel grid and direct mesh predictions can result in surface artifacts such as wrinkles and sharp protrusions. Body model parameter regression involves predicting the parameters of a statistical body model, such as SMPL \cite{SMPL:2015}, which provides a useful prior over body shape. Several approaches first predict a proxy representation from the input RGB image, such as surface keypoints \cite{Lassner:UP:2017, pavlakos2018humanshape}, silhouettes \cite{pavlakos2018humanshape, tan2017}, body part segmentations \cite{omran2018nbf} or IUV maps \cite{zhang2019danet, Xu_2019_ICCV}, and use this representation as the input to a regressor. Other approaches directly predict body model parameters from the input image \cite{hmrKanazawa17, kolotouros2019spin, Rong_2019_ICCV}. Fundamentally, learning-based approaches are dependent on the label accuracy and sample diversity of the training datasets used. This results in a significant drawback when the training data distribution is not sufficiently diverse in terms of body shape, pose and image (\eg background) conditions, as discussed below.

\noindent \textbf{3D human pose and shape datasets.} Learning-based approaches are trained using datasets of images paired with labels in the form of 3D joints or body model parameters. 3D labels may be obtained using motion capture (as for Human3.6M \cite{h36m_pami} and BML MoVi \cite{movi2020}), using inertial motion units (as for 3DPW \cite{vonMarcard2018}), or by optimisation (as for UP3D \cite{Lassner:UP:2017}). While current datasets contain varied and accurate 3D poses, they all suffer from limited body shape diversity, which greatly hampers the shape prediction accuracy of learning-based approaches. Additional drawbacks include: baggy clothing obscuring body shape and data captured in indoor MoCap environments being unrepresentative of in-the-wild images. We overcome these limitations of current training datasets by using synthetic training data. Furthermore, we create our own in-the-wild dataset, SSP-3D, to evaluate monocular 3D body shape predictions, which contains subjects with a greater variety of body shapes than current datasets.

\section{Method}
\label{sec:method}

\begin{figure}[t]
     \centering
     \begin{subfigure}[b]{0.73\textwidth}
         \centering
         \includegraphics[width=\textwidth]{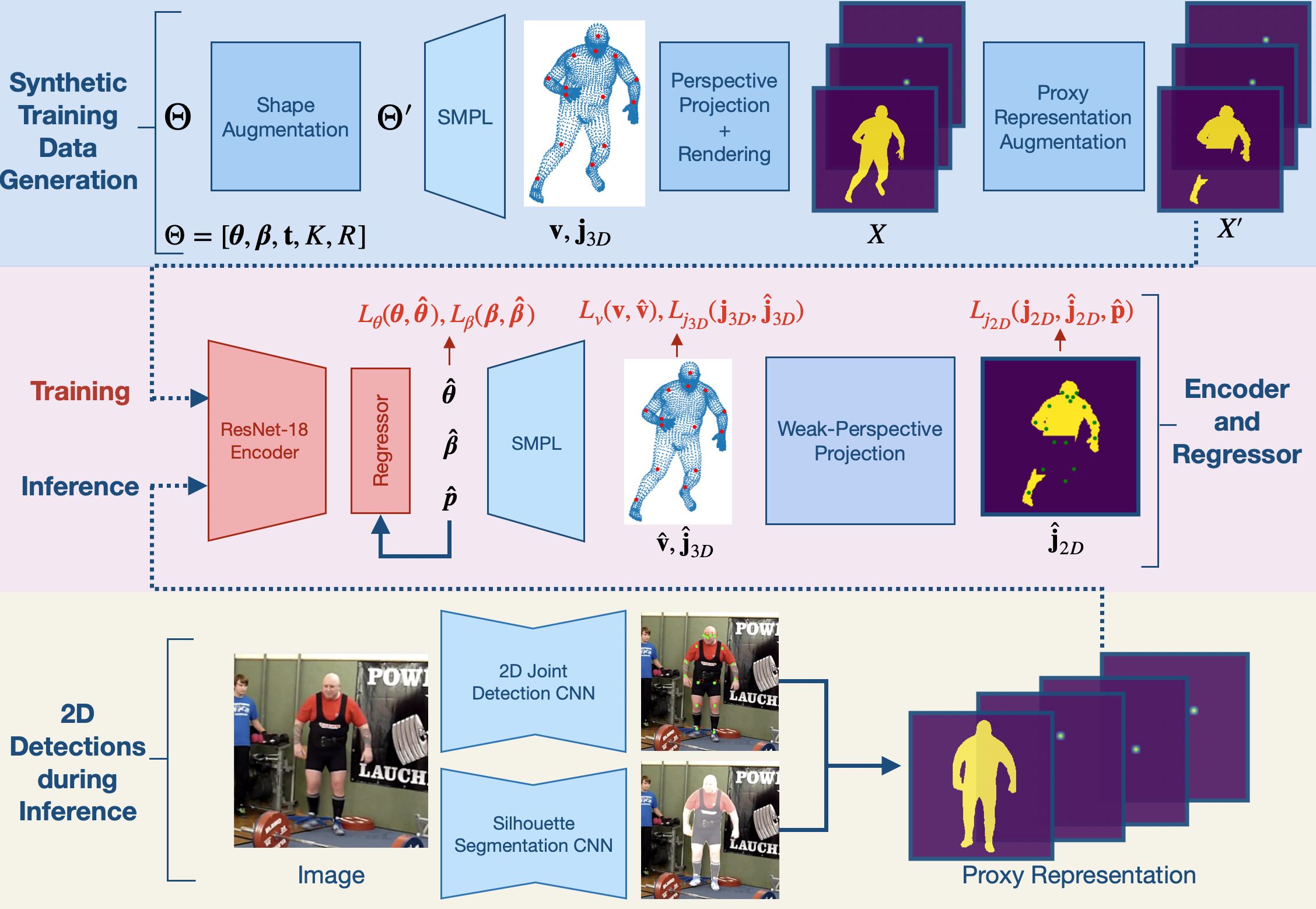}
         \caption{STRAPS Training and Inference}
         \label{fig:straps}
     \end{subfigure}
     \hfill
     \begin{subfigure}[b]{0.255\textwidth}
         \centering
         \includegraphics[width=\textwidth]{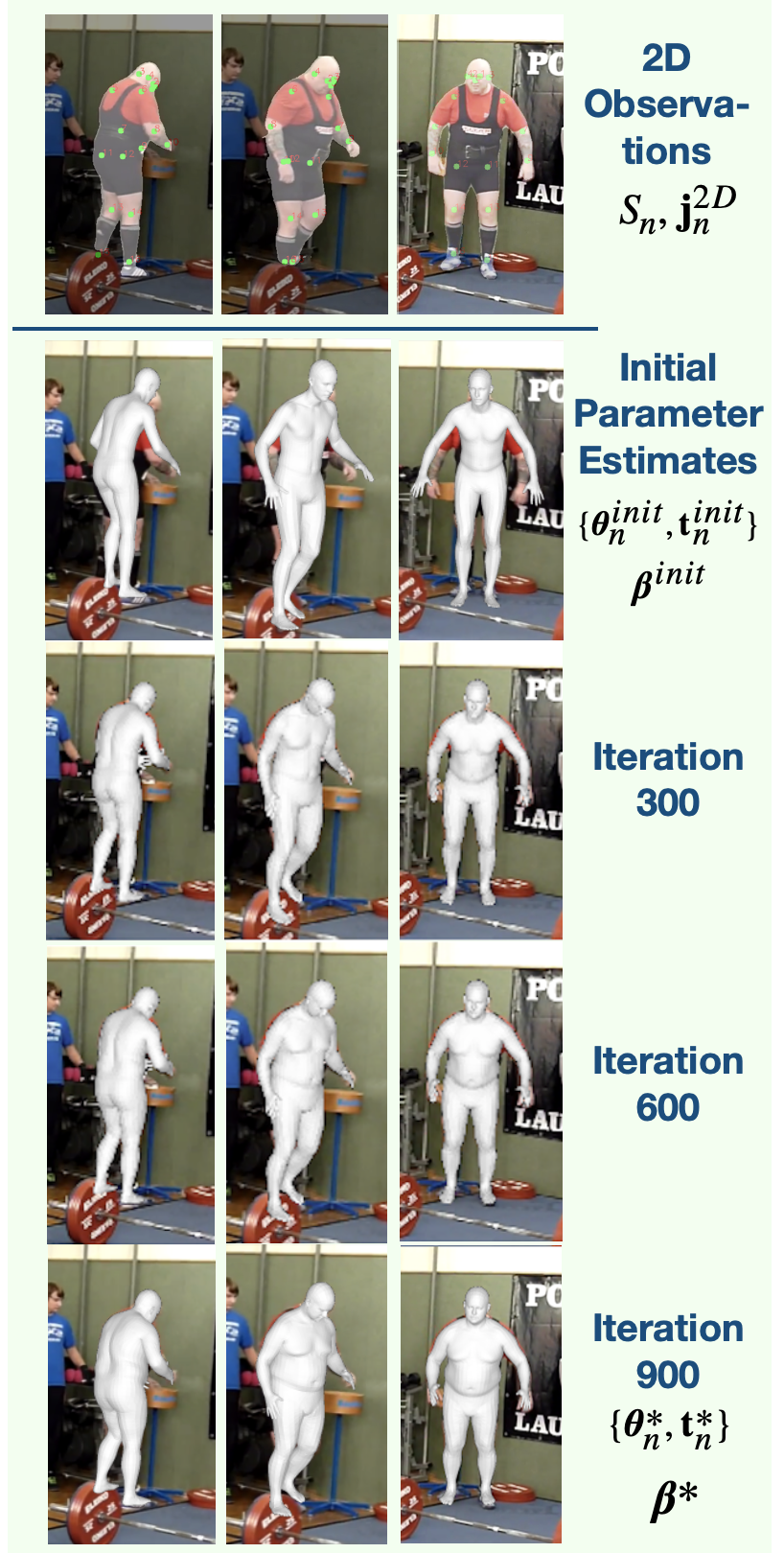}
         \caption{SSP-3D Optimisation}
         \label{fig:ssp_3d}
     \end{subfigure}
    \caption{\textbf{(a) Overview of the training and inference pipelines for STRAPS and (b) optimisation pipeline for SSP-3D.} Synthetic training data is generated by sampling SMPL \citep{SMPL:2015} pose and shape parameters and decoding them into 3D vertices and joints, which are projected, rendered and corrupted to form an input proxy representation. The proxy representation is passed through an encoder and iterative regressor (both with trainable weights) that predicts pose, shape and camera parameters. Supervision signals are applied to SMPL parameters, 3D joints and vertices, and 2D joints. At test-time, off-the-shelf detection and segmentation CNNs are used to create the input proxy representation. Optimisation for SSP-3D involves fitting SMPL to multi-frame 2D joints and silhouettes of a subject, which yields optimised pose and camera parameters for each frame and shape parameters for the subject.}
    \label{fig:combined_straps_ssp3d}
\end{figure}

In this section we describe STRAPS, our framework which utilises synthetic training data to overcome the lack of body shape diversity in real datasets. We also detail the multi-frame optimisation procedure used to create SSP-3D.

\subsection{STRAPS}
The proposed synthetic training process has two parts: synthetic data generation and neural encoder and regressor training, both of which use a parametric 3D body model.

\noindent \textbf{Parametric 3D body model}. The SMPL \cite{SMPL:2015} body model provides a fully-differentiable function $\mathcal{M}(\boldsymbol{\theta}, \boldsymbol{\beta})$ that takes shape-space coefficients $\boldsymbol{\beta}$ and 3D joint rotations $\boldsymbol{\theta}$ as inputs and outputs a human vertex mesh $\mathbf{v} \in \mathbb{R}^{N\times3}$. 3D joint locations are obtained as a linear combination of the vertices, $\mathbf{j}^{3D} = J\mathbf{v}$, where $J \in \mathbb{R}^{L\times N}$ is a regression matrix for $L$ joints of interest.

\noindent \textbf{Synthetic data generation}. SMPL is used to generate training data on-the-fly (top of Figure \ref{fig:straps}). In each iteration of the training loop, $\boldsymbol{\beta}$ and $\boldsymbol{\theta}$ are sampled from any training dataset with SMPL parameters - paired images are not required. A camera translation vector $\mathbf{t}$ is sampled randomly, while camera intrinsics and rotation matrices, $K$ and $R$, are fixed. To combat the insufficient body shape diversity in prevalent training datasets, we perform \textit{body shape augmentation} by replacing $\boldsymbol{\beta}$ with a new random vector $\boldsymbol{\beta}'$, generated by sampling each shape parameter $\beta'_n \sim \mathcal{N}(\mu, \sigma_n^2)$, where $\sigma_n$ is chosen (empirically) to provide greater body shape variance than current datasets. Then, the 3D vertices $\mathbf{v}$ and 3D joints $\mathbf{j}^{3D}$ corresponding to $\boldsymbol{\theta}$ and $\boldsymbol{\beta}'$ are perspective-projected and rendered \cite{kato2018renderer} into a silhouette $S \in [0, 1]^{H \times W} $ and 2D joint locations $\mathbf{j}^{2D} \in \mathbb{R}^{L \times 2}$. $\mathbf{j}^{2D}$ is transformed into 2D Gaussian joint heatmaps, $G \in \mathbb{R}^{H \times W \times L}$, where each channel corresponds to a separate joint location. We obtain our clean synthetic proxy representation (PR), $X \in \mathbb{R}^{H \times W \times (L+1)}$ by concatenating $S$ and $G$ along the channel dimension. Note that we opt for simple silhouettes and 2D joints as our PR, instead of more complex part segmentations or IUV maps \cite{Guler2018DensePose}, because the synthetic-to-real domain gap is smaller for a simple representation, and can be more easily bridged with \textit{proxy representation augmentation} during training. This involves modelling the failure modes of the off-the-shelf detection and segmentation CNNs used at test-time. In particular, noisy keypoint and silhouette predictions are modelled by adding uniform random noise to the 2D joint centres and silhouette edges in $X$. Occlusion is modelled by randomly removing body parts from and adding occluding boxes to the silhouette in $X$. The augmented PR $X'$ serves as the training input to our neural encoder. The training labels consist of $\boldsymbol{\theta}, \boldsymbol{\beta}', \mathbf{v}, \mathbf{j}_{3D}$ and $\mathbf{j}_{2D}$.

\noindent \textbf{Neural encoder and regressor.} STRAPS is architecture-agnostic. For this paper, we use the same network architecture as \cite{hmrKanazawa17, kolotouros2019spin}, which consists of a convolutional encoder for feature extraction and an iterative regressor that outputs predicted SMPL pose, shape and camera parameters ($\boldsymbol{\hat\theta}, \boldsymbol{\hat\beta}$ and $\mathbf{\hat p}$) given these features. Note that \cite{hmrKanazawa17, kolotouros2019spin} implement additional modules, namely an adversarial prior \cite{hmrKanazawa17} or “in-the-loop” optimisation \cite{kolotouros2019spin}, necessitated by their use of training images with only 2D joint labels. However, 2D joints crucially fail to supervise 3D shape, unlike our strong 3D supervision, which also does not require such additional modules. Weak-perspective camera parameters are predicted during regression, represented by $\mathbf{\hat p} = [\hat s, \mathbf{\hat t}]$, where $s \in \mathbb{R}$ represents scale and $\mathbf{\hat t} \in \mathbb{R}^2$ represents $x$-$y$ camera translation. We use a continuous 6-dimensional rotation representation for $\boldsymbol{\hat\theta}$ , as proposed by \cite{Zhou_2019_CVPR}, instead of the discontinuous Euler rotation vectors used by SMPL as default. From $\boldsymbol{\hat\theta}$ and $\boldsymbol{\hat\beta}$, SMPL is used to obtain predicted 3D vertices and joints, $\mathbf{\hat v}$ and $\mathbf{\hat j}^{3D}$. Finally, projected 2D joint predictions are obtained by $\mathbf{\hat j}^{2D} = s \Pi(\mathbf{\hat j}^{3D} + \mathbf{\hat t})$, where $\Pi$ represents an orthographic projection.

We train our network using a combination of 5 loss functions in a highly-multi-task framework. We use homoscedastic uncertainty \cite{kendall2017multi} to adaptively learn the loss weights during training, which results in an objective function of the form
\vspace{-0.25cm}
\begin{equation}
    L = \frac{1}{\sigma_\beta^2}L_{\beta} + \frac{1}{\sigma_\theta^2}L_{\theta} + \frac{1}{\sigma_{v}^2} L_{v} +  \frac{1}{\sigma_{j_{3D}}^2} L_{j_{3D}} +  \frac{1}{\sigma_{j_{2D}}^2} L_{j_{2D}} + \log(\sigma_\beta \sigma_\theta \sigma_v \sigma_{j_{3D}} \sigma_{j_{2D}}),
\vspace{-0.25cm}
\end{equation}
where the $\sigma^2$ terms represent task uncertainties and the $L$ terms represent mean squared error losses. Empirically, we found that redundancy in the multi-task objective - \eg applying losses on SMPL parameters as well as on 3D vertices, despite $\mathbf{v}$ being fully determined by $(\boldsymbol{\theta}, \boldsymbol{\beta})$ - improved both network convergence and final performance. We hypothesise that this is because each supervision signal has a different granularity. For instance, a loss on vertices provides a finer-scale supervision signal than a loss on 3D joints. Thus, we apply mean squared error losses on 3D joints ($L_{j_{3D}}$), 3D vertices ($L_v$), SMPL pose parameters in the 6D rotation representation of \cite{Zhou_2019_CVPR} ($L_\theta$), and SMPL shape coefficients ($L_\beta$). We employ an additional loss on projected 2D joints ($L_{j_{2D}}$) to enforce image-model alignment. 

\subsection{SSP-3D}

\begin{figure}[t]
     \centering
      \begin{subfigure}[c]{0.62\textwidth}
         \centering
         \includegraphics[width=\textwidth]{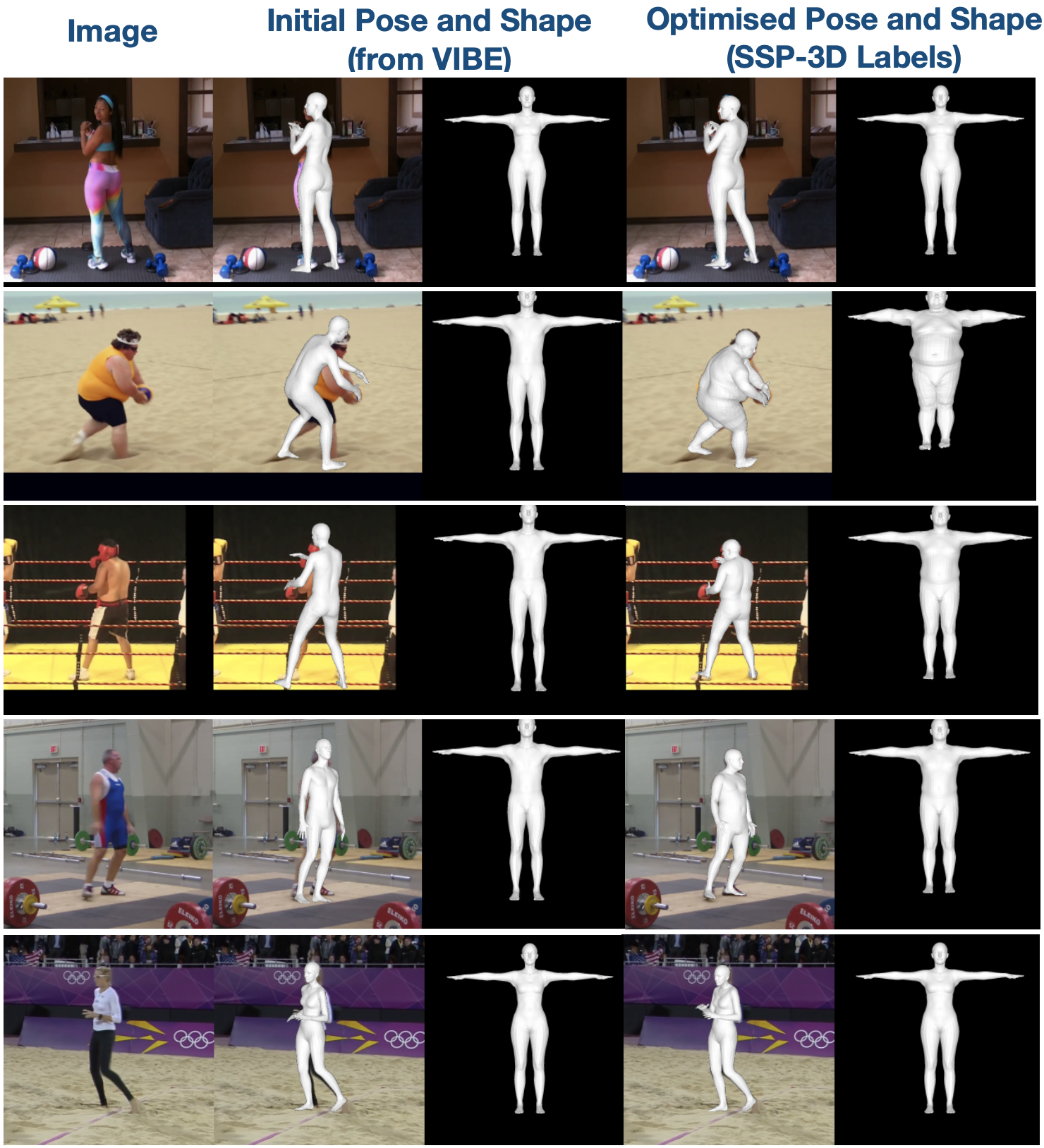}
         \vspace{-0.5cm}
         \caption{}
         \label{fig:ssp3d_pre_post_optimisation_samples}
         \vspace{0.05cm}
     \end{subfigure}
     \hfill
     \begin{subfigure}[c]{0.37\textwidth}
         \begin{subfigure}[c]{\textwidth}
             \centering
             \includegraphics[width=\textwidth]{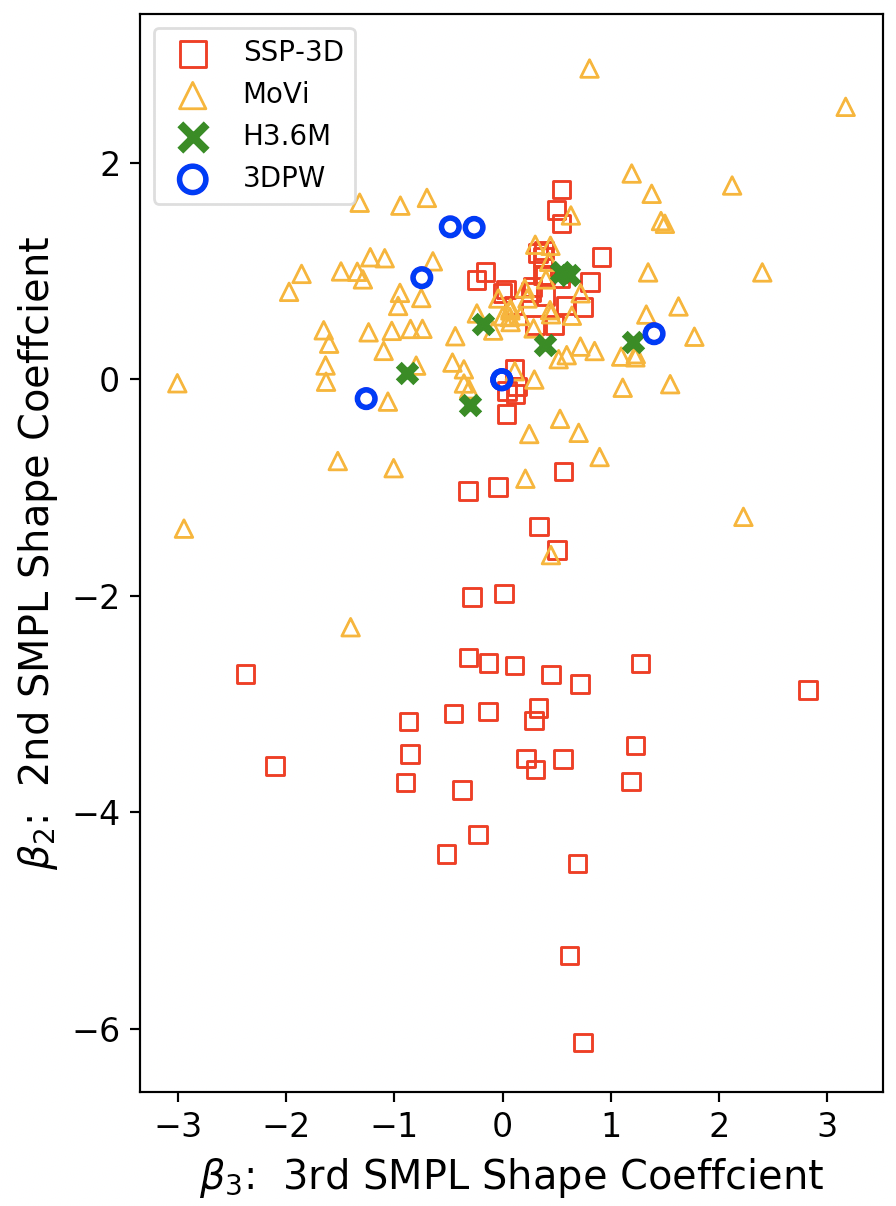}
             \vspace{-0.6cm}
             \caption{}
             \label{fig:datasets_beta_scatter}
             \vspace{0.05cm}
         \end{subfigure}
         \begin{subfigure}[c]{\textwidth}
            \centering
            \renewcommand{\tabcolsep}{2.0pt}
            \scriptsize
            \begin{tabular}{l|c|c|c}
            \hline
            \noalign{\smallskip} 
            Dataset & No. of Subjects & $\beta_2$ Var & $\beta_3$ Var\\
            \noalign{\smallskip}
            \hline
            \noalign{\smallskip}
            SSP-3D & 62 & 4.44 & 0.52\\
            H3.6M \cite{h36m_pami}& 7 & 0.17 & 0.41\\
            3DPW \cite{vonMarcard2018} & 7 & 0.39 & 0.49\\
            MoVi \cite{movi2020} & 86 & 0.68 & 1.40\\
            \noalign{\smallskip}
            \hline
            \end{tabular}
            \vspace{-0.1cm}
            \caption{}
            \label{table:datasets_beta_var}
         \end{subfigure}
        \end{subfigure}
    \vspace{-0.4cm}
    \caption{\textbf{SSP-3D samples and statistics.} (a) shows RGB images and corresponding optimised SMPL body shape and pose labels provided in the SSP-3D dataset. It also illustrates the improvement in shape and pose parameter estimates after optimisation, compared to initial estimates obtained with VIBE \cite{kocabas2019vibe}. (b) and (c) demonstrate the greater body shape diversity in SSP-3D compared to widely-used datasets, by considering the distribution of the 2nd and 3rd shape coefficient labels $\beta_2$ and $\beta_3$. (b) plots $\beta_2$ and $\beta_3$ for each sample in each dataset. (c) gives the number of subjects and the variance of $\beta_2$ and $\beta_3$ labels in each dataset. Note that $\beta_2$ is strongly correlated with variation in body fat content. $\beta_1$ is not used because it is strongly correlated with overall size, which is ambiguous in monocular predictions.}
    \label{fig:ssp3d_samples_statistics}
\end{figure}

SSP-3D contains 311 in-the-wild images of 62 tightly-clothed sports-persons (selected from the Sports-1M video dataset \cite{KarpathyCVPR14}) with a diverse range of body shapes, along with corresponding pseudo-ground-truth SMPL shape and pose labels. Figure \ref{fig:ssp3d_samples_statistics} illustrates the greater body shape diversity in SSP-3D compared to Human3.6M \cite{h36m_pami}, 3DPW \cite{vonMarcard2018} and MoVi \cite{movi2020}. Note that Human3.6M and MoVi are not in-the-wild and have homogeneous backgrounds.

SMPL shape and pose labels were acquired via optimisation, using an extended version of SMPLify \cite{Bogo:ECCV:2016} in a similar manner to the UP-3D dataset \cite{Lassner:UP:2017}. Unlike \cite{Bogo:ECCV:2016} and \cite{Lassner:UP:2017}, we used multiple frames of the same subject in parallel, obtaining different optimised poses $\boldsymbol{\theta}_n^*$ and camera translations $\mathbf{t}_n^*$ for each frame $n$, but \textit{forcing the optimised shape} $\boldsymbol{\beta}^*$ \textit{to be the same across all frames} to exploit multi-view information. We used 2D joints, acquired using Keypoint-RCNN \cite{he2017maskrcnn}, and pixel-accurate silhouettes, acquired using PointRend \cite{kirillov2019pointrend} on top of FPN \cite{lin2016fpn}, as the target 2D observations into which we fit the SMPL model. The use of multi-frame silhouettes ensures that SSP-3D has significantly more accurate body shape labels than UP-3D. To prevent getting stuck in bad local minima, we obtained an initialisation for per-frame pose and camera parameters, $\boldsymbol{\theta}_n^\text{init}$ and $\mathbf{t}_n^\text{init}$, and shape coefficients, $\boldsymbol{\beta}^\text{init}$, by using VIBE \cite{kocabas2019vibe}, a method for SMPL prediction from video (see Figure \ref{fig:ssp3d_pre_post_optimisation_samples}). Suitable frames for optimisation, with good SMPL initialisations and accurate target silhouettes and 2D joints, were hand-picked by human annotators. Our objective function is the sum of 6 error terms:
\vspace{-0.25cm}
\begin{equation}
    E(\boldsymbol{\beta}, \{\boldsymbol{\theta}_n, \mathbf{t}_n\}_{n=1}^N) =  \lambda_j E_j + \lambda_S E_S + \lambda_a E_a + \lambda_{\theta} E_\theta + \lambda_{\beta} E_\beta + \lambda_{\theta^\text{init}} E_{\theta^\text{init}},
\label{eqn:objective_func}
\vspace{-0.25cm}
\end{equation}
where $N$ is the number of frames and the $\lambda$ terms represent weights. The silhouette error term, $E_S$, penalises the $L_1$ difference between target and SMPL silhouettes. It is defined as
\vspace{-0.25cm}
\begin{equation}
    E_S(\boldsymbol{\beta}, \{\boldsymbol{\theta}_n, \mathbf{t}_n\}_{n=1}^N;  \{S_n\}_{n=1}^N) = \frac{1}{N} \sum_{n=1}^N \frac{\| \hat{S}(\Pi_K(\hat{\mathbf{v}}_n(\boldsymbol{\beta}, \boldsymbol{\theta}_n) + \mathbf{t}_n)) - S_n \|_1}{WH},
\vspace{-0.25cm}
\end{equation}
where $W, H$ are the width and height of the target silhouettes $S_n$ and $\hat{\mathbf{v}}_n(\boldsymbol{\beta}, \boldsymbol{\theta}_n)$ represents the SMPL vertices for the $n$-th frame. $\Pi_K()$ is a perspective-projection with intrinsic camera parameters $K$. Neural Mesh Renderer \cite{kato2018renderer} is used to differentiably render SMPL silhouettes $\hat{S}_n$ from projected vertices.

$E_{\theta^\text{init}}$ is a pose regularisation term, which penalises the $L_2$ distance between the current and initial estimates (from VIBE \cite{kocabas2019vibe}) of the SMPL pose parameters in rotation matrix form. We observed that the optimiser would use perspective effects to fit the SMPL model to target silhouettes of large persons, instead of updating the shape parameters. For example, the global rotation parameters would be updated to make the SMPL body lean towards the camera, enlarging the rendered silhouette. $E_{\theta^\text{init}}$ was incorporated to prevent such effects. It is defined as 
\vspace{-0.25cm}
\begin{equation}
    E_{\theta^\text{init}}(\{ \boldsymbol{\theta}_n\}_{n=1}^N ; \{\boldsymbol{\theta^\text{init}}_n\}_{n=1}^N) = \frac{1}{N} \sum_{n=1}^N \frac{\|\mathbf{r}(\boldsymbol{\theta}_n) - \mathbf{r}(\boldsymbol{\theta}_n^\text{init}) \|_2^2}{|\mathbf{r}(\boldsymbol{\theta}_n)|},
\vspace{-0.25cm}
\end{equation}
where $\mathbf{r}(\boldsymbol{\theta}_n) \in \mathbb{R}^{216}$ represents the vector of flattened and concatenated rotation matrices (for each of the 24 SMPL joints) corresponding to $\boldsymbol{\theta}_n$. 

$E_j$, $E_a$, $E_\theta$ and $E_\beta$ are derived from SMPLify and full definitions can be found in \cite{Bogo:ECCV:2016}. In short, $E_j$ is a weighted 2D joint reprojection error, $E_a$ is an angle prior term which penalises unnatural bending of the elbow and knees, $E_\theta$ is the negative log-likelihood of a Gaussian mixture model pose prior and $E_\beta$ is a $L_2$ regularisation penalty upon shape parameters. 

We optimise our objective function using the Adam \cite{kingma2014adam} optimiser with a learning rate of 0.01. After convergence, a human annotator selects good SMPL fits. Details on the human annotation, as well as all hyperparameter values, are available in the supplementary material.

\section{Implementation Details}
\textbf{Training datasets.} To generate synthetic training data, we sample SMPL pose parameters from the training sets of UP-3D \cite{Lassner:UP:2017} and 3DPW \cite{vonMarcard2018}, and from Human3.6M \cite{h36m_pami} subjects S1, S5, S6, S7 and S8 (after applying MoSh \cite{Loper:SIGASIA:2014} to obtain SMPL poses from 3D joint labels). For our baseline experiments without shape augmentation, we also use the SMPL shape parameters from these datasets. Synthetic silhouettes are cropped and resized to $256 \times 256$.

\noindent \textbf{Evaluation datasets.} We report metrics on Human3.6M (Protocol 2 \cite{hmrKanazawa17} subjects S9, S11), 3DPW (test set), BML-MoVi \cite{movi2020} (F-PG1 videos) and SSP-3D. For Human3.6M and 3DPW, we report mean per joint position error after rigid alignment with Procrustes analysis (MPJPE-PA \cite{mono-3dhp2017}). For BML-MoVi and SSP-3D, we report scale-corrected per-vertex Euclidean error in a neutral pose (or T-pose), i.e. PVE-T-SC. A description of the scale-correction technique used to combat scale ambiguity is given in the supplementary material. We also report silhouette mean intersection-over-union on SSP-3D.

\noindent \textbf{Architecture.} We use a ResNet-18 \cite{He2015} encoder, the output of which is average pooled, producing a feature vector $\boldsymbol{\phi} \in \mathbb{R}^{512}$. The iterative regression network consists of two fully connected layers with 512 neurons each, followed by an output layer with 157 neurons. We use the Adam \cite{kingma2014adam} optimiser to train our encoder and regressor, with a learning rate of 0.0001 and a batch size of 140. We train for 240 epochs, which takes 5 days on a single 2080Ti GPU. During inference, 2D joint predictions are obtained using Keypoint-RCNN \cite{he2017maskrcnn} and silhouette predictions are obtained using DensePose \cite{Guler2018DensePose}. All implementations are in PyTorch \cite{Pytorch_NEURIPS2019_9015}. Inference runs at $\sim$4fps, 90\% of which is silhouette and joint prediction.

\section{Empirical Evaluation}
\label{sec:empirical_results}

In this section, we present results from our ablative study, which investigates the effects of shape and proxy representation augmentation during synthetic training. We also compare our method to other approaches in terms of shape and pose accuracy.

\begin{figure}[t]
     \centering
     \begin{subfigure}[c]{0.49\textwidth}
         \centering
         \includegraphics[width=\textwidth]{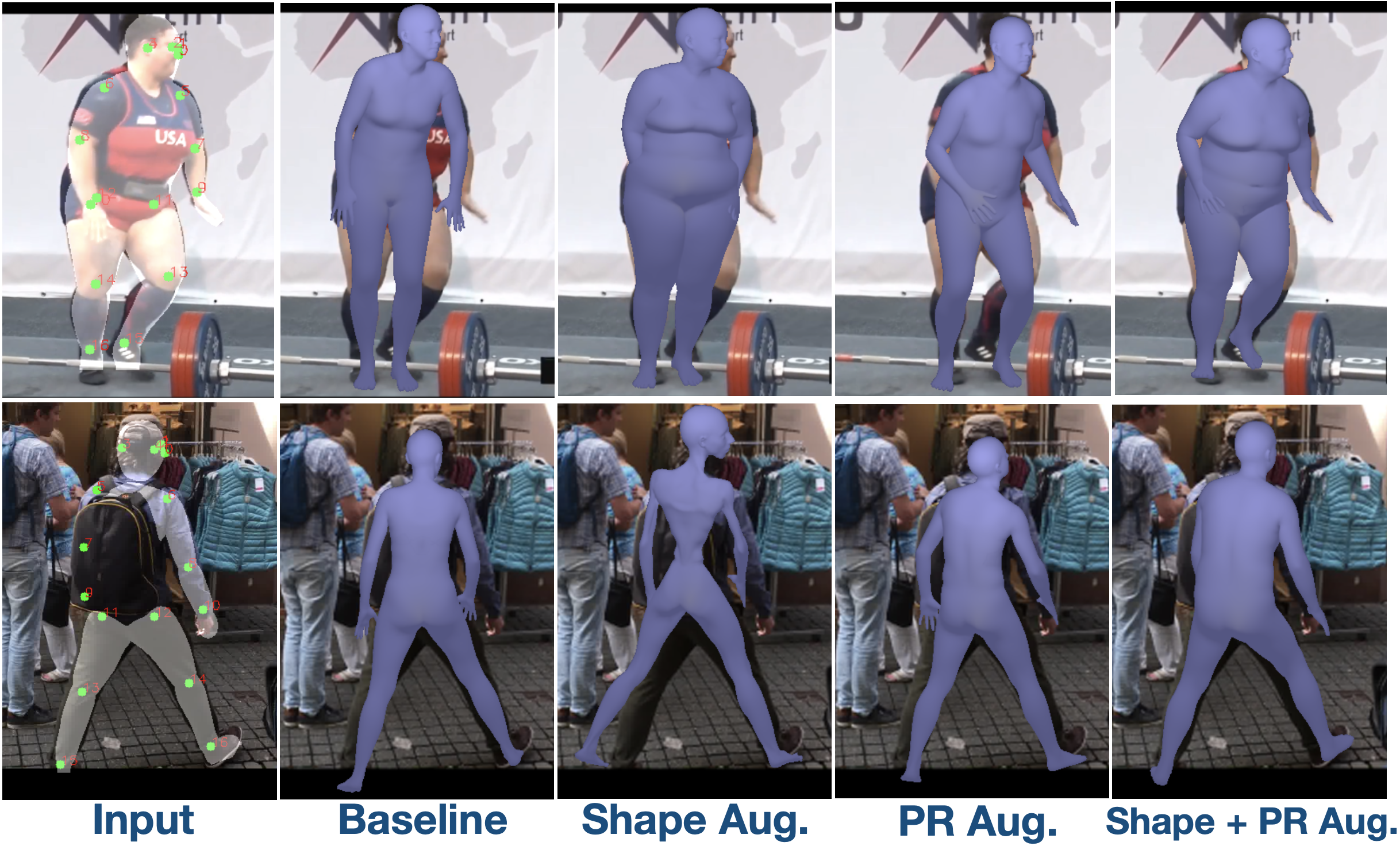}
         \vspace{-0.7cm}
         \caption{}
         \label{fig:ablation_fig}
     \end{subfigure}
     \hfill
     \begin{subfigure}[c]{0.49\textwidth}
        \renewcommand{\tabcolsep}{1.3pt}
        \scriptsize
        \begin{tabular}{l|l|c|c|c}
        \hline
        \noalign{\smallskip} 
        \multirow{2}{3em}{Input} & \multirow{2}{1em}{Augmentation} & SSP-3D & H3.6M & 3DPW\\
        & & PVE-T-SC & MPJPE-PA & MPJPE-PA\\
        \noalign{\smallskip}
        \hline
        \noalign{\smallskip}
        \multirow{4}{4em}{GT Synthetic} & Baseline & 14.4 & 40.4 & 39.0\\
        & Shape aug. & 10.1 & 34.1 & \textbf{34.9}\\
        & PR aug. & 16.1 & 42.2 & 44.7\\
        & Shape+PR aug. & \textbf{10.0} & \textbf{33.1} & 37.6\\
        \noalign{\smallskip}
        \hline
        \noalign{\smallskip}
        \multirow{4}{3em}{DP + KPRCNN} & Baseline & 20.1 & 70.5 & 71.3\\
        & Shape aug. & 24.1 & 75.5 & 88.6\\
        & PR aug. & 18.9 & 61.0 & 69.9\\
        & Shape+PR aug. & \textbf{15.9} & \textbf{55.4} & \textbf{66.8}\\
        \noalign{\smallskip}
        \hline
        \end{tabular}
        \vspace{-0.2cm}
        \caption{}
        \label{table:ablation_table}
     \end{subfigure}
    \vspace{-0.3cm}
    \caption{\textbf{Ablation study.} (a) illustrates that applying shape and proxy representation (PR) augmentation improves predictions of non-typical body shapes and develops robustness against noisy inputs. (b) reports pose (MPJPE-PA in mm) and shape (PVE-T-SC in mm) metrics when using synthetic proxy representations (rendered from ground-truth SMPL labels) versus predicted proxy representations (from DensePose and Keypoint-RCNN) as inputs.}
    \label{fig:ablation_composite}
\end{figure}

\begin{figure}[!h]
     \centering
     \begin{subfigure}[c]{0.44\textwidth}
         \centering
         \includegraphics[width=\textwidth]{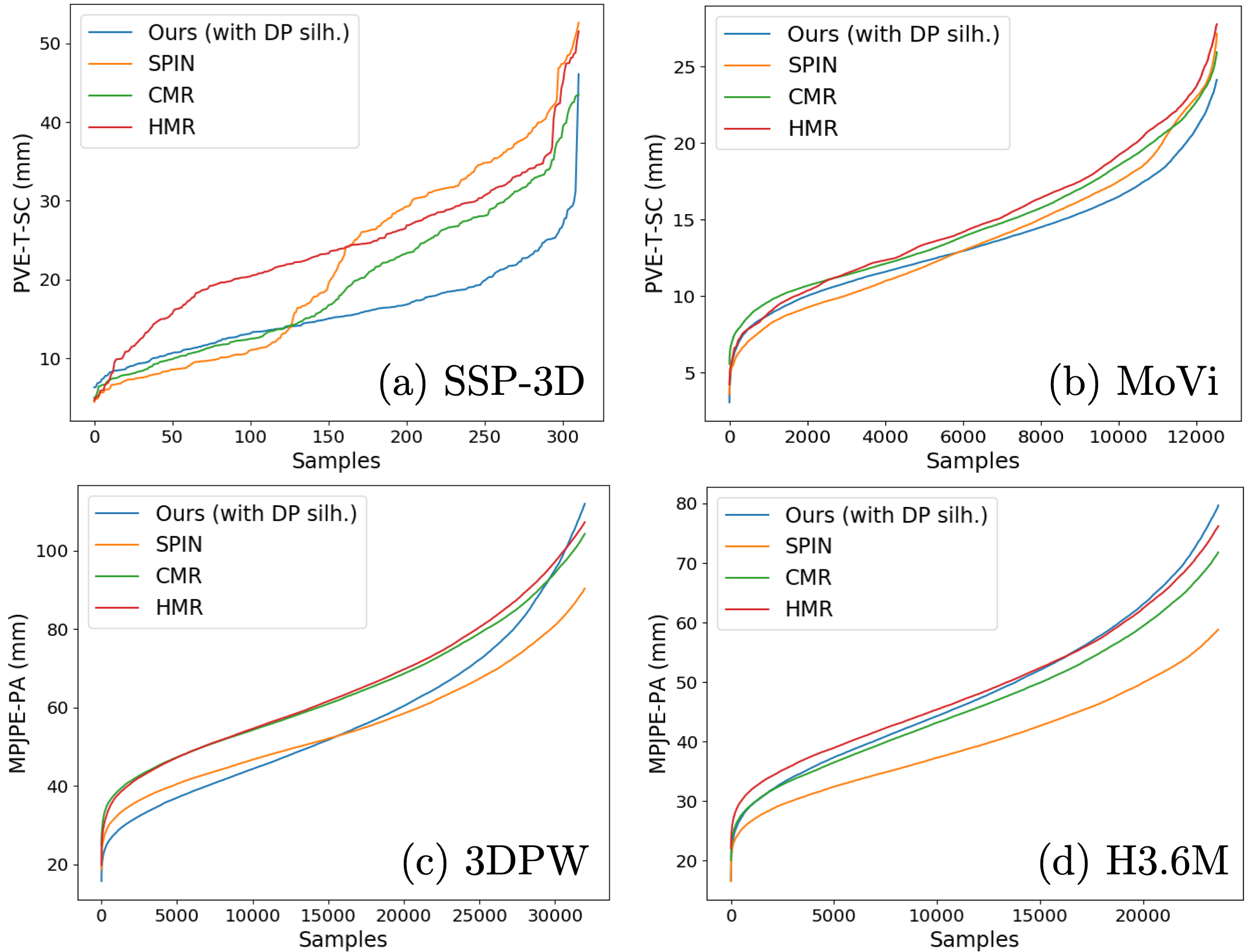}
         \vspace{-0.7cm}
         \caption{}
         \label{fig:error_dist_sota_comparison}
     \end{subfigure}
     \hfill
     \begin{subfigure}[c]{0.55\textwidth}
        \renewcommand{\tabcolsep}{1.3pt}
        \scriptsize
        \resizebox{\columnwidth}{!}{
        \begin{tabular}{l|cc|c|c|c}
        \hline
        \noalign{\smallskip}
        \multirow{2}{1em}{Method} & \multicolumn{2}{c|}{SSP-3D} & MoVi & H3.6M & 3DPW\\
        & PVE-T-SC & mIOU & PVE-T-SC & MPJPE-PA & MPJPE-PA\\
        \noalign{\smallskip}
        \hline
        \noalign{\smallskip}
        Ours (baseline) & 20.1 & 0.62 & \textbf{13.2} & 70.5 & 71.3\\
        Pavlakos \etal \cite{pavlakos2018humanshape} & - & - & - & 75.9 & -\\
        HMR (unpaired) \cite{hmrKanazawa17} & 20.8* & 0.61* & 14.2* & 66.5 & -\\
        SPIN (unpaired) \cite{kolotouros2019spin} & - & - & - & 62.0 & -\\
        Ours & \textbf{15.9} & \textbf{0.80} & \textbf{14.0} & 55.4 & 66.8\\
        \noalign{\smallskip}
        \hline
        \noalign{\smallskip}
        HMR \cite{hmrKanazawa17} & 22.9* & 0.69* & 15.5* & 56.8 & 71.5*\\
        NBF \cite{omran2018nbf} & 20.9* & - & 14.4* & 59.9 & 90.7*\\
        CMR \cite{kolotouros2019cmr} & 19.5* & 0.68* & 15.2* & 50.1 & 70.3*\\
        SPIN \cite{kolotouros2019spin} & 22.2* & 0.70* & 14.3* & \textbf{41.1} & \textbf{59.2}\\
        \noalign{\smallskip}
        \hline
        \end{tabular}
        }
        \vspace{-0.2cm}
        \caption{}
        \label{table:ssp3d_h36m_3dpw_movi}
     \end{subfigure}
    \vspace{-0.3cm}
    \caption{\textbf{Quantitative comparison with the SOTA} on SSP-3D, MoVi, Human3.6M (Protocol 2) and 3DPW. We report PVE-T-SC (mm) and mIOU on shape-centric datasets SSP-3D and MoVi and MPJPE-PA (mm) on pose-centric datasets 3DPW and Human3.6M. (a) plots the (sorted) distributions of metrics per evaluation sample. (b) lists mean metrics over all samples. Methods in the top part of (b) do not require training data comprised of images paired with 3D ground truth, while methods in the bottom part do. Numbers marked with * were evaluated for this paper, all other numbers are reported by the respective papers.}
    \label{fig:sota_comparison_composite}
\end{figure}

\noindent \textbf{Ablation studies.} Our ablative study investigates the effects of shape and proxy representation (PR) augmentation applied during synthetic training. We compare four networks, trained with: (i) no augmentation (baseline), (ii) only shape augmentation, (iii) only PR augmentation and (iv) shape + PR augmentation. Evaluations are carried out with two types of input proxy representations: synthetic silhouettes and 2D joints generated from GT SMPL labels and "real" silhouettes and 2D joints predicted from test RGB images using DensePose \cite{Guler2018DensePose} and Keypoint-RCNN \cite{he2017maskrcnn} respectively. SSP-3D evaluates 3D shape prediction across a diverse range of body shapes, while Human 3.6M and 3DPW evaluate 3D pose prediction.

The quantitative performance of the baseline network on GT synthetic inputs (see Figure \ref{table:ablation_table} first row) motivates the use of synthetic training data, since, in this ideal case, it achieves greater than SOTA accuracy (compare with Figure \ref{table:ssp3d_h36m_3dpw_movi}). However, in the practically-applicable situation using "real" inputs, the baseline network has two key failure modes: firstly, the predicted body shape is inaccurate, particularly for non-typical subjects (see Figure \ref{fig:ablation_fig}, top row) and secondly, the network becomes reliant on the perfectly-rendered synthetic inputs and is unable to deal with noisy real inputs. Incorporating shape augmentation alleviates the first problem, since the network sees a greater variety of shapes during training. However, the second problem is greatly exacerbated, particularly in cases with occluded silhouettes (see Figure \ref{fig:ablation_fig}, bottom row). Hence, the network trained with shape augmentation results in better shape (and pose) metrics on synthetic inputs compared to the baseline, as shown in Figure \ref{table:ablation_table}, while the metrics on real inputs are poor. Incorporating PR augmentation shrinks the performance deterioration when using real versus synthetic inputs by explicitly modelling input noise and occlusion during the synthetic training process. By combining PR and shape augmentation, we are able to predict a diverse range of body shapes, improve our pose accuracy significantly over the baseline and produce semantically-plausible outputs on all datasets, even when the input is heavily corrupted (see Figure \ref{fig:ablation_fig}).

\begin{figure}[!t]
    \centering
    \includegraphics[width=\textwidth]{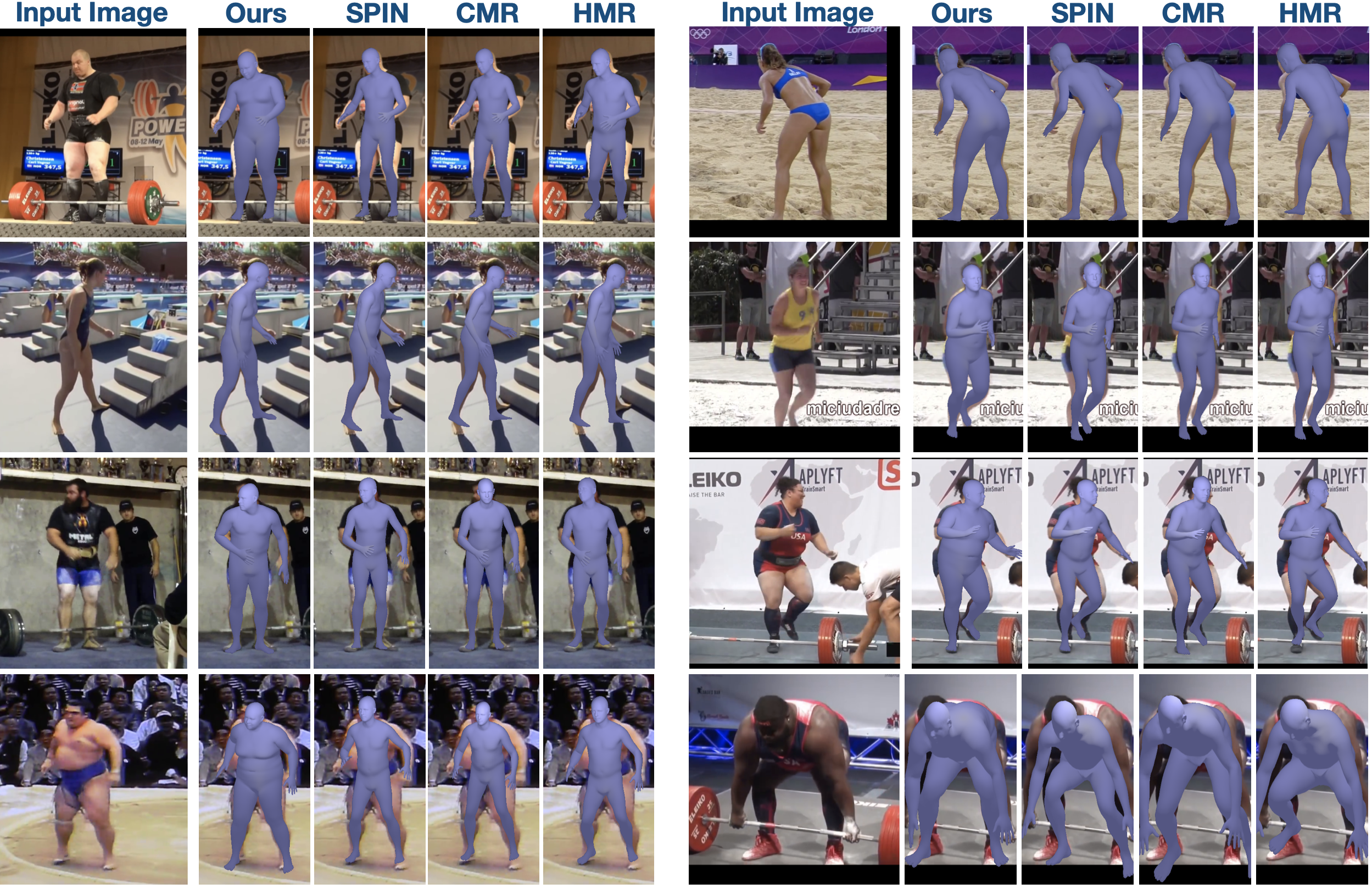}
    \caption{\textbf{Qualitative comparison on SSP-3D.} Each row shows examples from different PVE-T-SC quartiles (for our method), top to bottom: 0-25\%, 25-50\%, 50-75\%, 75-100\%. Results from SPIN \cite{kolotouros2019spin}, CMR \cite{kolotouros2019cmr} and HMR \cite{hmrKanazawa17} are shown for comparison. Our method is able to accurately predict a diverse range of body shapes, whereas other approaches are biased towards an average body shape prediction.}
    \label{fig:ssp3d_qualitative}
    \vspace{-0.2cm}
\end{figure}

\noindent \textbf{Comparison with the state-of-the-art.} Our method, with shape and PR augmentation, surpasses the state-of-the-art in terms of PVE-T-SC and mIOU on SSP-3D and MoVi. The distribution of errors per SSP-3D sample, shown in Figure \ref{fig:error_dist_sota_comparison}, suggests that our method is able to maintain shape prediction accuracy for challenging evaluation samples while the performance of competing approaches degrades for samples featuring non-average body shapes, qualitative examples of which are given in Figure \ref{fig:ssp3d_qualitative}. Our method may give erroneous reconstructions for outlier body shapes, in which case DensePose fails to predict an accurate silhouette, or due to poses with substantial self-occlusion, as shown in Figure \ref{fig:ssp3d_qualitative} bottom row.

\begin{figure}
    \centering
    \includegraphics[width=\textwidth]{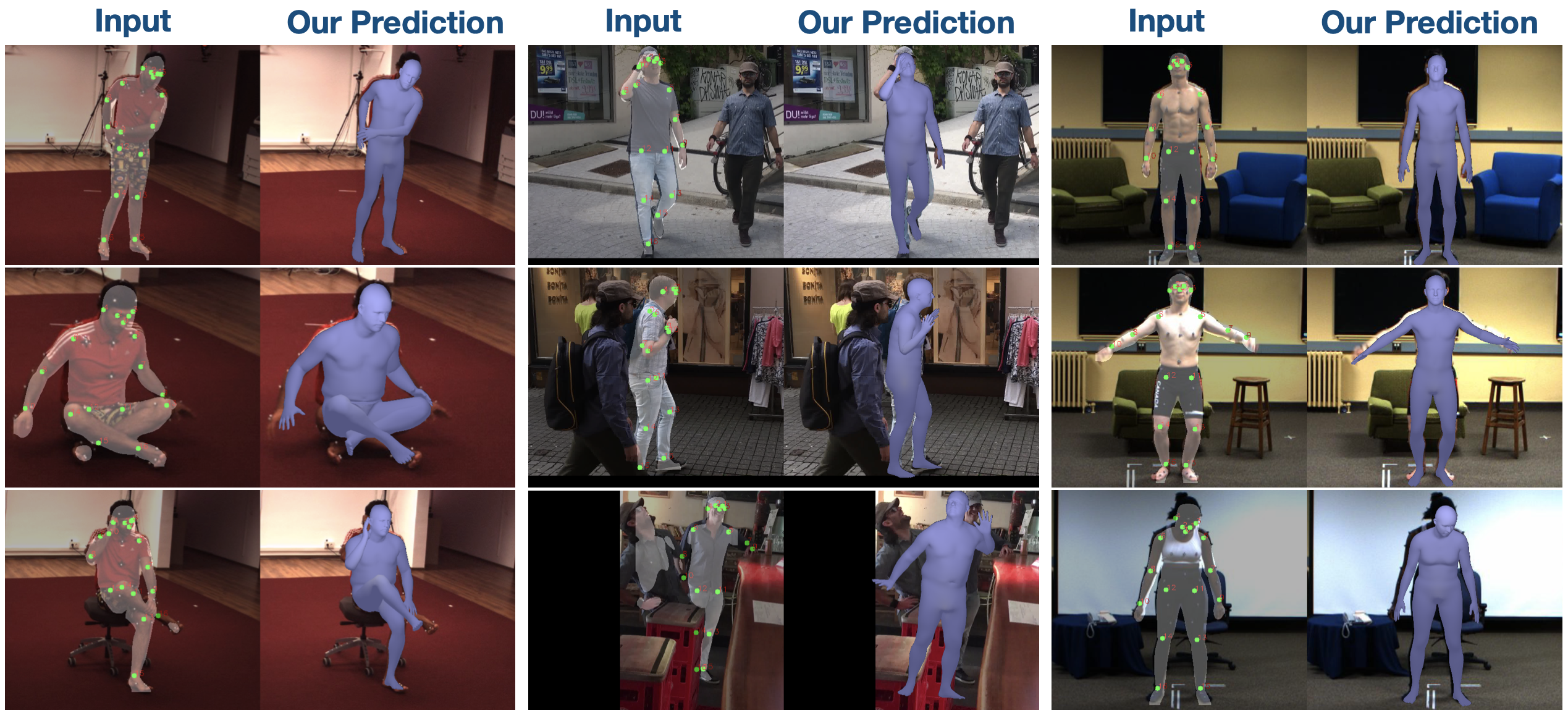}
    \caption{\textbf{Qualitative results on Human3.6M (left), 3DPW (middle) and MoVi (right).} Each row shows examples from different error metric quantiles (MPJPE-PA for H3.6M and 3DPW, PVE-T-SC for MoVi). Top to bottom: 0-33\%, 33-66\%, 66-100\%. Input silhouettes and input 2D keypoints are visualised over each image.}
    \label{fig:h36m_3dpw_movi_qualitative}
    \vspace{-0.2cm}
\end{figure}

Although we focus on shape prediction, our method is competitive with the SOTA on H3.6M and 3DPW in terms of MPJPE-PA and outperforms other methods that do not require training data comprised of images paired with expensive-to-obtain 3D labels. Qualitative examples are given in Figure \ref{fig:h36m_3dpw_movi_qualitative}. We observe that we perform relatively better on 3DPW than H3.6M, as compared to other methods, particularly up to the median error (Figure \ref{fig:error_dist_sota_comparison}). This is because methods trained on images captured in an indoor MoCap environment (like H3.6M) do not maintain the same pose prediction accuracy for test images with unconstrained background and lighting conditions (like in 3DPW). We create our input proxy representation using 2D segmentation and detection CNNs, which are more easily trained to be invariant to such variables. Thus, we match or surpass the SOTA on 3DPW for samples up to the median MPJPE-PA. However, 3DPW contains samples with severe occlusion (beyond what is modelled by PR augmentation) and overlapping persons, which cause DensePose to predict erroneous silhouettes and results in worse MPJPE-PA in the 75-100\% quartile.

\section{Conclusion}
In this paper, we addressed the problem of monocular 3D human shape and pose estimation. In particular, we observed that current approaches often predict inaccurate body shapes, particularly for non-typical subjects, due to a lack of body shape diversity in prevalent 3D human datasets. Thus, we proposed STRAPS, a learning framework that overcomes the lack of diversity by generating synthetic training data with diverse body shapes on-the-fly, such that the regressor sees a new body shape at every training iteration. To evaluate our approach, we created a challenging evaluation dataset for monocular human shape estimation, SSP-3D, which consists of RGB images of tightly-clothed sports-persons with a variety of body shapes and corresponding pseudo-ground-truth SMPL shape and pose parameters. We showed that STRAPS outperforms other approaches on SSP-3D in terms of shape prediction accuracy, while remaining competitive with the state-of-the-art on pose-centric datasets.

\clearpage
\bibliography{bib}
\end{document}